\documentclass[runningheads]{llncs}
\usepackage{graphicx}

\usepackage{tikz}
\usepackage{comment}
\usepackage{amsmath,amssymb} 
\usepackage{color}
\usepackage{xspace}

\usepackage{multirow}
\usepackage[noend]{algpseudocode}

\usepackage{algorithmicx,algorithm}
\usepackage{booktabs}
\usepackage{subfig}
\usepackage{hyperref}

\usepackage[accsupp]{axessibility}  

\makeatletter
\newcommand\onedot{\futurelet\@let@token\@onedot}
\def\@onedot{\ifx\@let@token.\else.\null\fi\xspace}

\def\eg{\emph{e.g}\onedot} 
\def\ie{\emph{i.e}\onedot}

\def\etal{\emph{et al}\onedot}
\makeatother

\newcommand{\SPL}{CORES$^2$}

\begin{document}
\pagestyle{headings}
\mainmatter
\def\ECCVSubNumber{4684}  

\title{Centrality and Consistency: Two-Stage Clean Samples Identification for Learning with Instance-Dependent Noisy Labels} 

\titlerunning{ECCV-22 submission ID \ECCVSubNumber} 
\authorrunning{ECCV-22 submission ID \ECCVSubNumber} 
\author{Anonymous ECCV submission}
\institute{Paper ID \ECCVSubNumber}

\titlerunning{Two-Stage Clean
Samples Identification for Learning with
IDN}
%
\author{Ganlong Zhao\inst{1,2}\orcidID{0000-0002-1612-3641} \and
Guanbin Li\inst{1}\thanks{Corresponding authors are Guanbin Li and Yizhou Yu.}\orcidID{0000-0002-4805-0926} \and
Yipeng Qin\inst{3}\orcidID{0000-0002-1551-9126} \and
Feng Liu\inst{4}\orcidID{0000-0002-4811-7828} \and
\\ Yizhou Yu\inst{2*}\orcidID{0000-0002-0470-5548}}
\authorrunning{G. Zhao et al.}
%

\institute{Sun Yat-sen University, Guangzhou 510006, China \and
The University of Hong Kong, Hong Kong, China \and 
Cardiff University, Cardiff, United Kingdom \and
Deepwise AI Lab, Beijing, China \\
\email{zhaogl@connect.hku.hk, liguanbin@mail.sysu.edu.cn, QinY16@cardiff.ac.uk, liufeng@deepwise.com, yizhouy@acm.org}}

\maketitle

\begin{abstract}
Deep models trained with noisy labels are prone to over-fitting and struggle in generalization. Most existing solutions are based on an ideal assumption that the label noise is class-conditional, \ie~instances of the same class share the same noise model, and are independent of features. While in practice, the real-world noise patterns are usually more fine-grained as instance-dependent ones, which poses a big challenge, especially in the presence of inter-class imbalance. In this paper, we propose a two-stage clean samples identification method to address the aforementioned challenge. 
First, we employ a class-level feature clustering procedure for the early identification of clean samples that are near the class-wise prediction centers. Notably, we address the class imbalance problem by aggregating rare classes according to their prediction entropy. Second, for the remaining clean samples that are close to the ground truth class boundary (usually mixed with the samples with instance-dependent noises), we propose a novel consistency-based classification method that identifies them using the consistency of two classifier heads:~the higher the consistency, the larger the probability that a sample is clean. Extensive experiments on several challenging benchmarks demonstrate the superior performance of our method against the state-of-the-art. Code is available at \url{https://github.com/uitrbn/TSCSI\_IDN}.
\end{abstract}

\section{Introduction}

Deep learning has shown transformative power in various real-world applications but is notoriously data-hungry\cite{he2015delving,he2016deep,ren2015faster,he2017mask,long2015fully,NMI21}.
There are some other alternatives which try to reduce the cost of human labor for data annotation, such as crawling web images and using machine-generated labels. 
However, such data are usually noisy, which impedes the generalization of deep learning models due to over-fitting.

Addressing the aforementioned issue, Learning with Noisy Labels (LNL) was proposed as a new topic and has attracted increasing attention in both academia and industry.
Existing LNL methods mostly focus on the learning with class-conditional noise~(CCN), which aims to recover a noise transition matrix that contains class-dependent probabilities of a clean label flipping into a noisy label.
However, CCN is too ideal for real-world LNL as it ignores the dependence of noise on the content of individual images, {\it a.k.a.} instance-dependent noise~(IDN). 

\begin{figure}[t]
  \centering
   \includegraphics[width=0.4\linewidth]{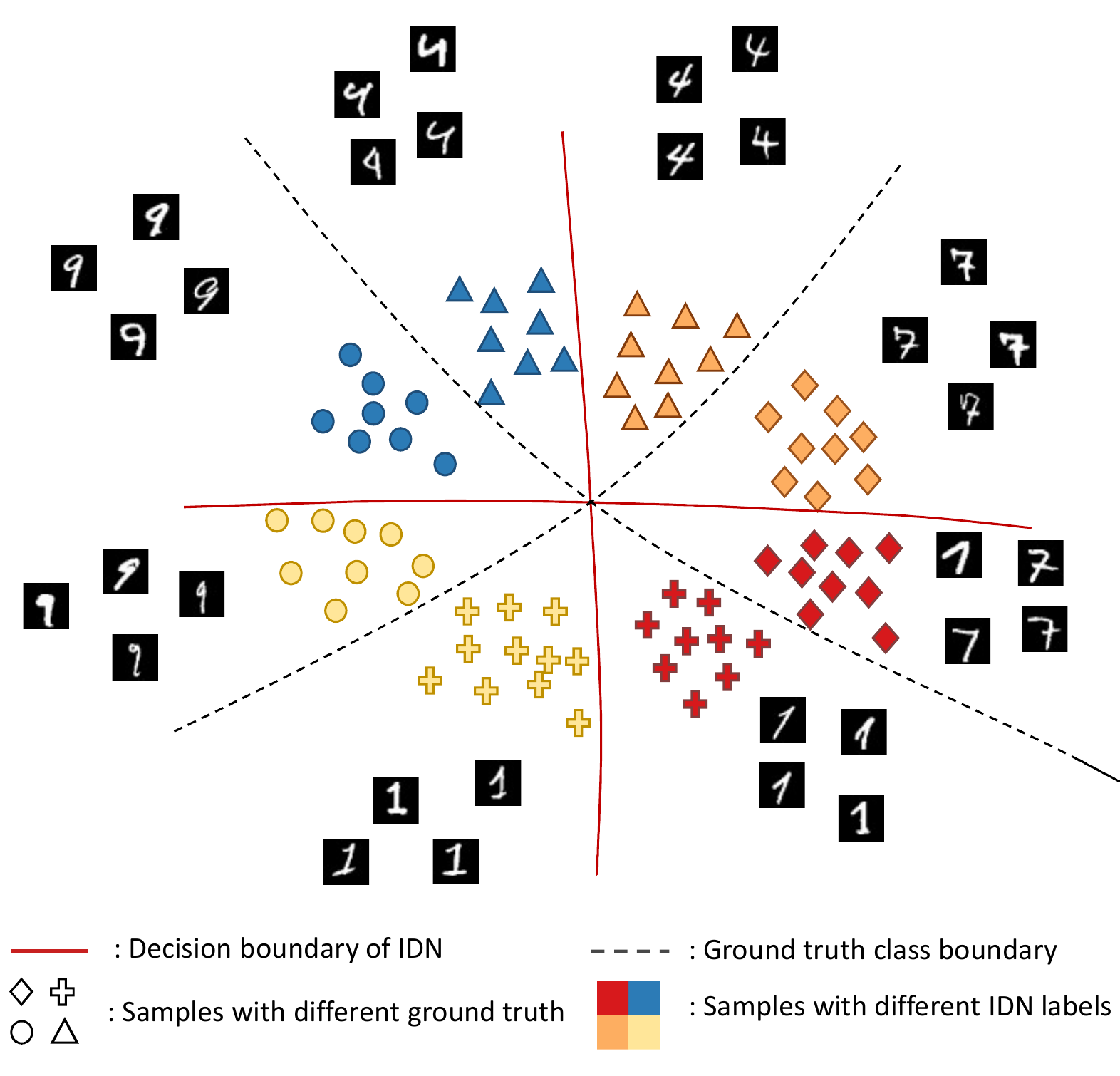}

   \caption{Example of IDN. The different shapes of the markers represent different ground truth classes. The different colors of the markers represent the noisy (IDN) labels. Different from random noise, IDN samples tend to be distributed near the ground truth class boundary, thus confusing the classifier and leading to over-fitted decision boundaries.}
   \label{fig:idn_example}
\end{figure}

Unlike random noise or CCN that can be countered by collecting more (noisy) data\cite{chen2020robustness}, IDN has some important characteristic that makes it difficult to be tackled.
First, classifiers can easily over-fit to the IDN because the noisy labels are dependent on sample features. 
As Fig.~\ref{fig:idn_example} shows, mislabeled IDN samples (samples with the same shape but with different colors) share similar image features to their mislabeled classes, and thus tend to be distributed near the boundary between their ground truth class and the mislabeled class. 
As a result, the classifier can easily be confused and over-fits to IDN samples, leading to specious decision boundaries (red lines in Fig.~\ref{fig:idn_example}).
In addition, the challenge of IDN can be further amplified in the presence of inter-class imbalance and differences. 
Consider Clothing1M~\cite{xiao2015learning}, an IDN dataset verified by~\cite{chen2020beyond}, in which the noise is highly imbalanced and asymmetric.
In Clothing1M, the IDN samples are unevenly distributed as the samples from similar classes (\eg sweater and knitwear) can be extremely ambiguous, while those from other classes (\eg shawl and underwear) are easily distinguishable.
Such unevenly distributed IDN samples can be further amplified by the class imbalance problem, as there is no guarantee of a balanced dataset due to the absence of ground truth labels.

\begin{figure}[t]
  \centering
   \includegraphics[width=0.6\linewidth]{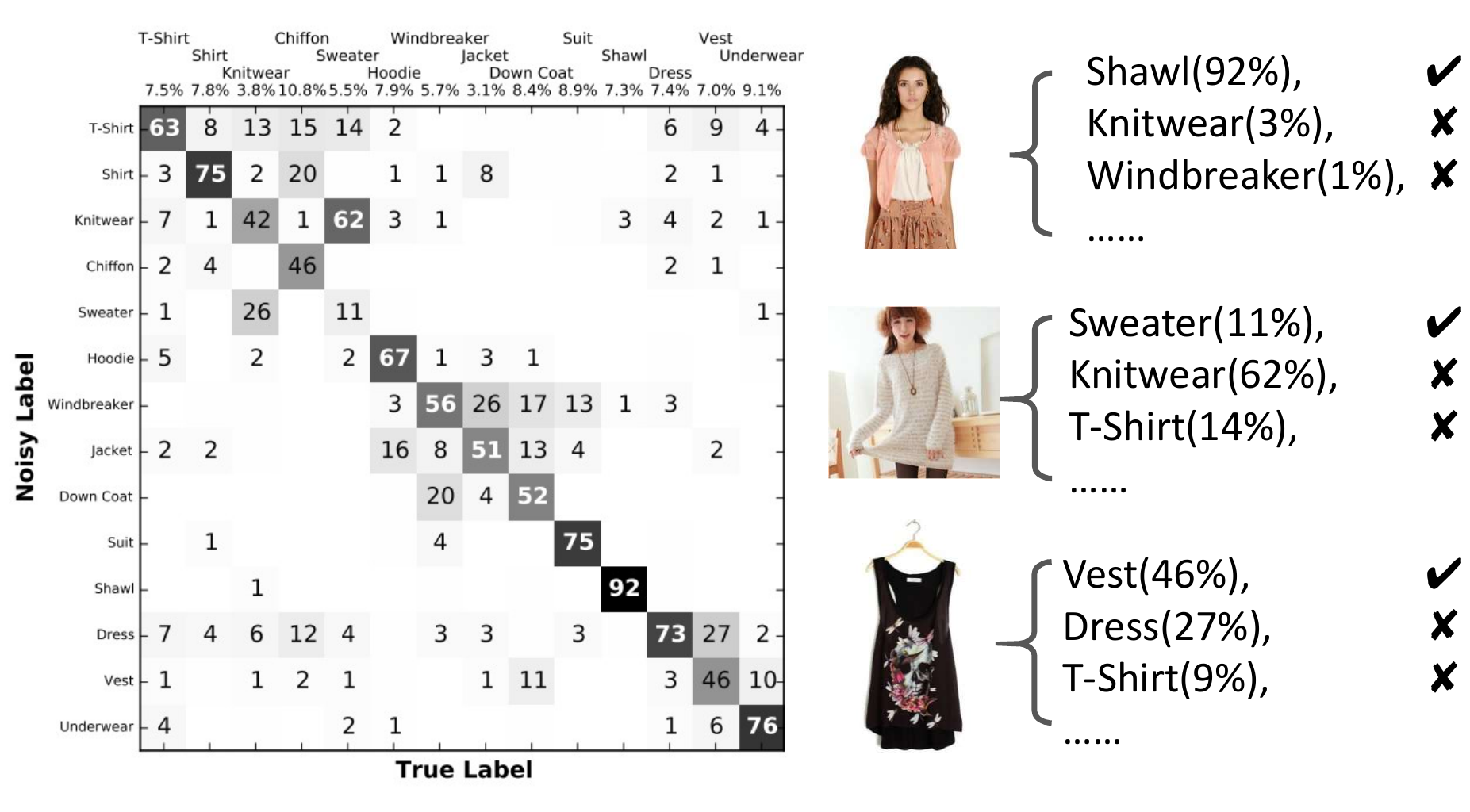}

   \caption{The transition matrix of Clothing1M copied from \cite{xiao2015learning}. The distribution of noisy labels are highly imbalanced. Some classes are almost clean (e.g. Shawl) while some classes has more mislabeled samples than correct labels (e.g. Sweater).}
   \label{fig:clothing1m_matrix}
\end{figure}

In this paper, we follow DivideMix~\cite{Li2020DivideMix} that formulates LNL as a semi-supervised learning problem and propose a novel two-stage method to identify clean versus noisy samples in the presence of IDN and the class imbalance problem.
In the first stage, we employ a class-level feature-based clustering procedure to identify easily distinguishable clean samples according to their cosine similarity to the corresponding class-wise prediction centers.
Specifically, we collect the normalized features of samples belonging to different classes respectively and calculate their class-wise centers located on a unit sphere. Then, we apply Gaussian Mixture Model (GMM) to binarily classify the samples according to their cosine similarity to their corresponding class centers and identify the ones closer to class centers as clean samples.
Notably, we propose to augment the GMM classification by aggregating rare classes based on their prediction entropy, thereby alleviating the impact of the class imbalance problem.
In the second stage, we propose a consistency-based classification method to identify the hard clean samples that are mixed with IDN samples around the ground truth class boundaries.
Our key insight is that such clean samples can be identified by the prediction consistency of two classifiers. Compared to IDN samples, clean samples should produce more consistent predictions.
Specifically, we incorporate two regularizers into the training: one applied to the feature extractor to encourage it to facilitate consistent outputs of the two classifiers; one applied to the two classifiers to enforce them generating inconsistent predictions.
After training, we use another GMM to binarily classify the samples 
with smaller GMM means as clean samples. 
After identifying all clean samples, we feed them into the semi-supervised training as labeled samples, thereby implementing our learning with instance-dependent noisy labels.
In summary, our contributions could be summarized as:
\begin{itemize}
    \item We propose a method that delving into the instance-dependent noise, and design a class-level feature clustering procedure focusing on the imbalanced and IDN samples detection.
    \item We further propose to 
    identify the hard clean samples around the ground truth class boundaries by measuring the prediction consistency between two in-dependently trained classifiers, and further improves the accuracy of clean versus noisy classification.
    \item Our method achieves state-of-the-art performance in some challenging benchmarks, and is proved to be effective in different kinds of synthetic IDN.
\end{itemize}

\section{Related Work}

A large proportion of previous LNL methods focus on the class-conditional noise. With the class-conditional noise assumption, some methods try to correct the loss function with the noise transition matrix\cite{patrini2017making}, which can be estimated through exploiting a noisy dataset\cite{liu2015classification,patrini2017making,xia2020extended,zhu2021clusterability} or using a clean set of data\cite{hendrycks2018using,zheng2021meta}.
Such loss correction methods based on noise transition matrix is infeasible for instance-dependent noise, since the matrix is dataset dependent and the number of parameters grows proportionally with the size of training dataset. 

Some methods seek to correct the loss by reweighting the noisy samples or selecting the clean data~\cite{thulasidasan2019combating,konstantinov2019robust}. A common solution is to treat the samples with smaller loss as clean data\cite{Li2020DivideMix,shen2019learning,jiang2018mentornet}. However, as pointed out by \cite{chen2020beyond}, instance-dependent noise can be more easily over-fitted, and the memorization effect, which indicates that CNN-based models always tend to learn the general simple pattern before over-fitting to the noisy labels, becomes less significant when the model is trained with instance-dependent noise.

Some other methods combat the noisy label with other techniques. For example, Kim \etal\cite{kim2021joint} combine positive learning with negative learning, which uses the complementary labels of noisy data for model training. Some methods\cite{Li2020DivideMix,nguyen2019self} formulate LNL as a semi-supervised learning problem. DivideMix\cite{Li2020DivideMix} divides the dataset into clean and noisy sets, which serve as labeled and unlabeled data for semi-supervised learning. Some methods investigate the influence of augmentation strategy\cite{nishi2021augmentation} or enforce the prediction consistency between different augmentations\cite{lu2021co}. C2D\cite{zheltonozhskii2021contrast} utilizes self-supervised learning to facilitate the learning with noisy labels.

Chen \etal\cite{cheng2020learning} pointed out that for diagonally-dominant class-conditional noise, one can always obtain an approximately optimal classifier by training with a sufficient number of noisy samples. And it raise the significance of learning with IDN. There has been some works for this topic. CORES$^2$\cite{cheng2020learning} try to progressively sieve out corrupted samples and avoid specifying noise rate. CAL\cite{zhu2021second} propose a second-order approach with the assistance of additional second-order statistics. Besides, some research work also propose methods for IDN generation\cite{chen2020beyond,xia2020parts}.

\begin{figure*}[t]
  \centering
   \includegraphics[width=1.0\linewidth]{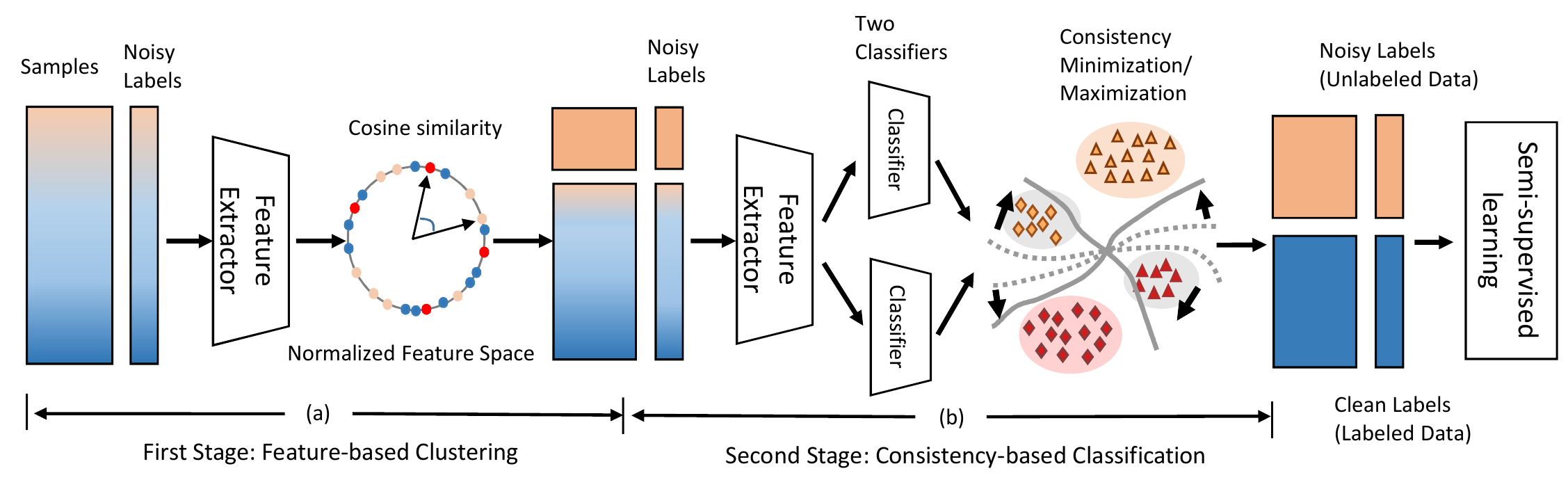}

   \caption{The overview of our proposed method. (a) The first stage. The noisy samples and labels are sent to the feature extractor for calculating the normalized features. The features are clustered with the prediction of samples. Noisy samples are divide to clean set and noisy set according to the cosine similarity between the feature and the center of its labels. (b) The model is train to minimize/maximize the prediction between two classifier heads and samples with smaller consistency are identified as noisy labels. (c) The clean/noisy set serve as labeled/unlabeled data for semi-supervised training.}
   \label{fig:overview}
\end{figure*}

\section{Method}

\subsection{Overview}

The classification of noisy versus clean samples by the model outputs and their labels is a prevalent choice in the learning with noisy labels (LNL). Previous studies use the cross-entropy of noisy samples~\cite{Li2020DivideMix} or confidence thresholds~\cite{yao2021jo} for noisy versus clean division. 
However, as Chen \etal~\cite{chen2020beyond} point out, samples with instance-dependent noise (IDN) can be more easily over-fitted by neural networks, resulting in less reliable model outputs that confuse the classification of clean versus noisy samples.
Such confusion is further amplified when the noisy dataset is imbalanced.
For example, the differences between clean and noisy samples might be neglected for rare classes that contribute little to the overall prediction accuracy. 

Therefore, we propose a two-stage method which can effectively address IDN in the presence of class imbalance.
In the first stage, we leverage a class-level feature-based clustering process to identify easily distinguishable clean samples that are close to their corresponding class centers in the feature space. Specifically, in this stage, we address the class imbalance by aggregating rare classes identified by their prediction entropy.
In the second stage, we address the remaining clean samples, which are close to the ground truth class boundaries and are thus mixed with IDN samples.
Our key insight is that such clean samples can be identified by the consistent predictions of two classifiers.
Specifically, we propose a mini-max strategy for this consistency-based clean versus noisy classification: we simultaneously regularize the two classifiers to generate inconsistent predictions but enforce the feature extractor to facilitate the two classifiers to generate consistent predictions.
After training, we identify the clean samples as the ones that lead to more consistent predictions between the two classifiers. 
After identifying all clean samples, we follow DivideMix~\cite{Li2020DivideMix} and implement the learning with instance-dependent noisy labels as a semi-supervised learning problem that takes the clean samples as labeled samples, and the rest (noisy) samples as unlabeled samples. 

\subsection{Feature-based Clustering}
\label{sec:feature_clustering}

As common practice, we divide a CNN-based classifier into two parts: a feature extractor $F$ that takes images as input and extracts their features, and the following classifier $G$ that outputs classification probabilities based on the image features extracted by $F$. 
Given a noisy dataset $\{x_i, \bar{y_i}\}^{N}_{i=1}$, where $x_i$ is an image sample and $\bar{y_i}$ is its (noisy) label, we denote $\hat{f_i} = \frac{f_i}{\|f_i\|}$ as the normalized feature of $x_i$ extracted by $F$, \ie $f_i = F(x_i)$, $\hat{y_i}=G(f_i)$ as the predicted label of $x_i$, and calculate the class-wise feature centers $O_c$ according to $\hat{y_i}$ as:
\begin{equation}
    O_c = \frac{\sum_{i=1}^{N_c}{\hat{f_i}}}{\|\sum_{i=1}^{N_c}{\hat{f_i}}\|},
    \label{eq:class_centers}
\end{equation}
where $c \in \{1,2,3,...,C\}$ denotes the $C$ classes, $N_c$ is the number of samples $x_i$ whose noisy label $\bar{y_i} = c$. Then, we can obtain the cosine similarity between each sample $x_i$ and its corresponding feature center $O_{\bar{y_i}}$ as:
\begin{equation}
    S_i = \hat{f_i} \cdot O_{\bar{y_i}}.
    \label{eq:cos_sim}
\end{equation}
Finally, we apply class-wise Gaussian Mixture Model (GMM) to the similarities $S_i$ of samples for each class and performs binary classification. As the cosine similarity of noisy samples tend to be smaller, the component of GMM with a larger mean, \ie larger similarity, is denoted as the clean set. Thus all the noisy samples is classified as clean or noisy as the preliminary result of first stage.

\noindent \textbf{Entropy-based Aggregation of Rare Classes}
However, the performance of the proposed feature-based clustering can be unstable when the sizes of some classes are small and not sufficient for binary classification, which often happens in real-world datasets that have large numbers of classes.
Addressing this issue, we propose to aggregate rare classes that struggle with the proposed binary classification. 
Specifically, we set a class aggregate threshold $\theta_{agg}$ and calculate the average prediction entropy of the samples for each class $c$ as:
\begin{equation}
    \mathrm{Ent}(c) = -\frac{1}{N_c}\sum_{i=1}^{N_{c}} \sum_{j=1}^B p_i^j \log p_i^j,
\end{equation}
where $N_c$ is the number of samples for class $c$, $B=2$ indicates the binary classification of clean versus noisy samples, $p_i^j$ represents the output probability that a sample $x_i$ belongs to class $j$, \ie, clean and noisy probability.
Samples of class $c$ that satisfy $\mathrm{Ent}_{c} > \theta_{agg}$ are aggregated and treated as a single class to facilitate our feature-based clustering.

\subsection{Consistency-based Classification}
\label{sec:discrepancy_classification}

\begin{figure*}[t]
  \centering
   \includegraphics[width=\linewidth]{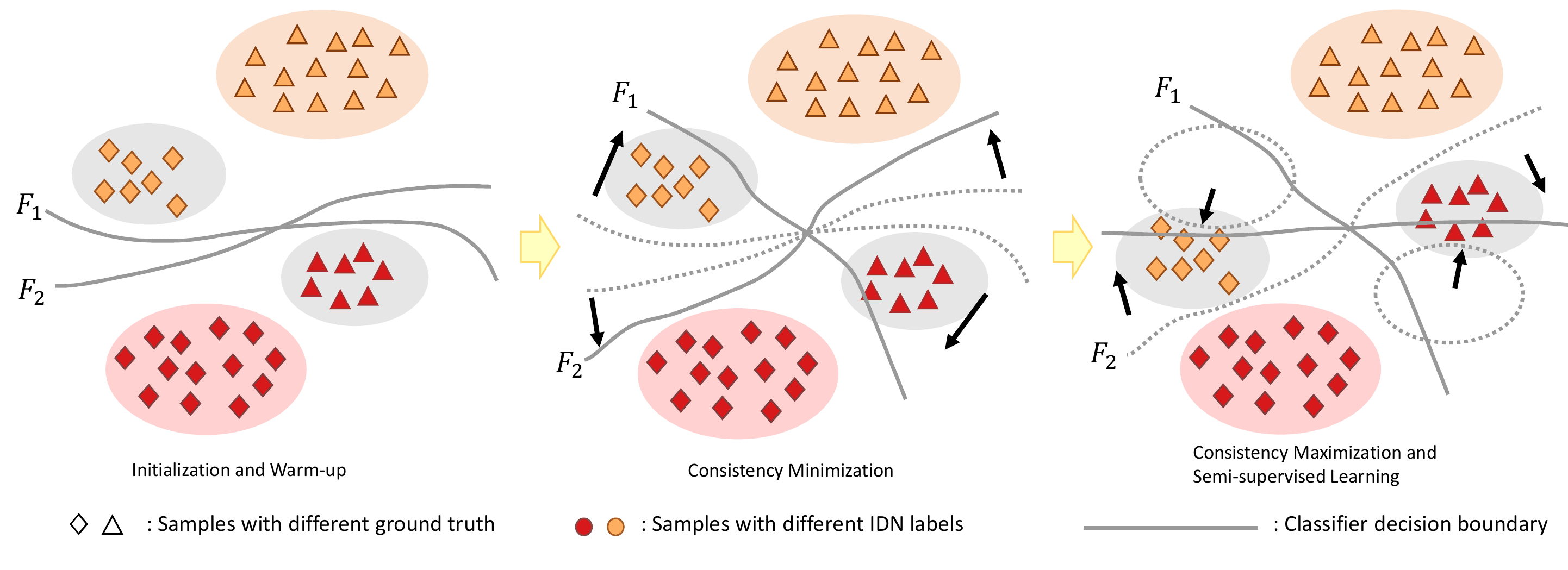}

   \caption{The procedure of the consistency-based classification. At the beginning, two classifiers has different prediction due to different initialization. Then the prediction consistency between two classifiers is minimized to identify the ambiguous noisy samples near the decision boundary. At the third steps, feature extractor is trained to maximize the consistency and the semi-supervised loss further revises both feature extractor and classifiers. }
   \label{fig:mcd_procedure}
\end{figure*}

As Fig~\ref{fig:idn_example} shows, challenging clean samples are usually near the ground truth class boundaries in the feature space, which can be identified by the consistency between two independently trained classifiers $G_1$ and $G_2$ that have different decision boundaries. Therefore, by replacing the classifier $G$ with $G_1$ and $G_2$ in our network, we can get two corresponding predictions $p_x^1$ and $p_x^2$ of the same sample $x$.
Then, we define and calculate the consistency between $G_1$ and $G_2$ on $x$ as:
\begin{equation}
    D(p^1, p^2) = \sum_{i=1}^{C}|p_i^1 - p_i^2|,
\end{equation}
where $x$ is omitted for simplicity and $C$ is the number of classes, \ie the dimension of $p_x^1$ and $p_x^2$. We measure the discrepancy with $L1$ norm following \cite{saito2018maximum}.

\noindent \textbf{Consistency Minimization Regularization}
Although being independently trained, $G_1$ and $G_2$ share the same training dataset and the same loss function, leading to a non-negligible risk that the corresponding two predictions are identical or very similar. 
To minimize such a risk, we propose to incorporate a regularization loss on $G_1$ and $G_2$ that aims to minimize their consistency:
\begin{equation}
    L_{\mathrm{min}} = -\lambda_{\min} \sum_{i=1}^{N} D^*(p_{x_i}^1, p_{x_i}^2),
    \label{eq:classifer_dis_loss}
\end{equation}
where $N$ is the number of samples and $\lambda_{\min}$ controls the strength,
\begin{equation}
    D^*(p_x^1, p_x^2) = w_{C_x}\sum_{i=1}^{C}|p_i^1 - p_i^2|,
\end{equation}
where $x$ is omitted on the right side for simplicity and $w_{C_x}$ is the frequency of samples $x$'s noisy category $C_x$. $w_{C_x}$ is used to counter the class imbalance problem that often happens in real-world datasets. As the GMM model in the first stage does not guarantee the inter-class balance in the clean set, $w_{C_x}$ explicitly increases the weight of classes with more samples in consistency minimization and thus more samples are filtered out.

\noindent \textbf{Consistency Maximization Regularization}
Solely using the minimization regularization might impair the model performance because the consistency of samples with correct labels are also minimized, and ideally two classifiers should output the same prediction for each sample. Therefore, we propose to add a consistency maximization loss on the feature extractor $F$ to constrain the network:
\begin{equation}
    L_{\mathrm{max}} = \lambda_{\max} \sum_{i=1}^{N} D^*(p_{x_i}^1, p_{x_i}^2),
    \label{eq:backbone_dis_loss}
\end{equation}
where $\lambda_{\max}$ controls the strength.
Furthermore, the maximization of $L_{\mathrm{max}}$ forces the feature extractor to separate the ambiguous features and thus complements semi-supervised training. As shown in the third step of Fig.~\ref{fig:mcd_procedure}, the feature extractor maximizes the consistency by pushing the samples with small consistency towards clean labeled data, and semi-supervised learning tries to gather the the feature of similar samples.

\subsection{Training Procedure}

Based on the discussions in Sec.~\ref{sec:feature_clustering} and Sec.~\ref{sec:discrepancy_classification}, we propose to train our model by repeating the following four steps for each epoch.

\noindent \textbf{Initialization} 
Before training, we following~\cite{Li2020DivideMix} and warm up our model including the two classifiers for several epochs with all noisy labels, where steps 1 and 2 belong to our feature-based clustering (Stage 1), and steps 3 and 4 belong to our consistency-based classification (Stage 2).

\noindent
\textbf{Step-1}
We first extract the features of noisy data and calculate the class-wise feature centers according to Eq.~\ref{eq:class_centers}. Then, we calculate the cosine similarities between features and the center of noisy labels of each sample using Eq.~\ref{eq:cos_sim}.

\noindent
\textbf{Step-2} 
We perform a binary (noisy vs. clean) classification to samples by applying class-wise Gaussian Mixture Model (GMM) according to the cosine similarities obtained in Step-1. 
We label the GMM component with a larger mean as ``clean''. Then, we select the samples with clean probabilities higher than a threshold $\theta$ as our primary clean set $S^1_{clean}$ and the rest samples as the noisy set $S^1_{noisy}$.

\noindent
\textbf{Step-3} 
We first fix the feature extractor and train the two classifiers to minimize their consistency according to Eq.~\ref{eq:classifer_dis_loss} for $N_{\max}$ iterations using $S^1_{clean}$. 
Then, we evaluate the consistency of all samples in $S^1_{clean}$. 
Similar to Step-2, we apply a GMM model to the consistencies and select the samples with small mean as clean set $S^2_{clean}$. The rest samples are merged with $S^1_{noisy}$ as $S^2_{noisy}$.

\noindent
\textbf{Step-4} 
With $S^2_{clean}$ and $S^2_{noisy}$ obtained as above, we optimize our model with a supervised loss on $S^2_{clean}$ and a semi-supervised loss on $S^2_{noisy}$:
\begin{equation}
L = L_\mathcal{X} + \lambda_\mathcal{U} L_\mathcal{U}
\end{equation} 
where $S^2_{clean}$ and $S^2_{noisy}$ are used as labeled set $\mathcal{X}$ and unlabeled set $\mathcal{U}$ respectively, and $\lambda_\mathcal{U}$ balances the trade-off between $L_\mathcal{X}$ and $L_\mathcal{U}$. In addition, we add additional consistency maximization regularization (Eq.~\ref{eq:backbone_dis_loss}) to the feature extractor during training.

\section{Experiment}

In this section, we will validate the effectiveness of our method on several benchmark datasets with different kinds of IDNs (\ie synthetic and real-world ones) and different numbers of classes.

\begin{table*}[!t]
		\caption{Comparison of test accuracies ($\%$) using different methods on CIFAR10 and CIFAR100 with part-dependent label noise. Results of other methods are copied from CAL\cite{zhu2021second}. Our method outperforms all previous methods in all settings.}
		\begin{center}
		\resizebox{0.8\columnwidth}{!}{%
		{\begin{tabular}{c|cccccc} 
				\hline 
				 \multirow{2}{*}{Method}  & \multicolumn{3}{c}{\emph{Inst. CIFAR10} } & \multicolumn{3}{c}{\emph{Inst. CIFAR100} } \\ 
				 & $\eta = 0.2$&$\eta = 0.4$&$\eta = 0.6$ & $\eta = 0.2$&$\eta = 0.4$&$\eta = 0.6$\\
				\hline\hline
			     CE (Standard)  &85.45$\pm$0.57 & 76.23$\pm$1.54 &  59.75$\pm$1.30 &  57.79$\pm$1.25 & 41.15$\pm$0.83 & 25.68$\pm$1.55 \\
				 Forward $T$ \cite{patrini2017making} & 87.22$\pm$1.60 & 79.37$\pm$2.72 & 66.56$\pm$4.90  & 58.19$\pm$1.37 & 42.80$\pm$1.01 & 27.91$\pm$3.35\\
				 $L_{\sf DMI}$ \cite{xu2019l_dmi}  &88.57$\pm$0.60 & 82.82$\pm$1.49 & 69.94$\pm$1.31 & 57.90$\pm$1.21 & 42.70$\pm$0.92 & 26.96$\pm$2.08\\
				 $L_{q}$ \cite{zhang2018generalized}   & 85.81$\pm$0.83 & 74.66$\pm$1.12 & 60.76$\pm$3.08  & 57.03$\pm$0.27 & 39.81$\pm$1.18 & 24.87$\pm$2.46\\
				 Co-teaching \cite{han2018co} & 88.87$\pm$0.24 & 73.00$\pm$1.24 & 62.51$\pm$1.98 & 43.30$\pm$0.39 & 23.21$\pm$0.57 & 12.58$\pm$0.51\\
				 Co-teaching+ \cite{yu2019does}  & 89.80$\pm$0.28 & 73.78$\pm$1.39 & 59.22$\pm$6.34 & 41.71$\pm$0.78 & 24.45$\pm$0.71 & 12.58$\pm$0.51\\
				JoCoR \cite{wei2020combating} & 88.78$\pm$0.15 & 71.64$\pm$3.09 & 63.46$\pm$1.58 & 43.66$\pm$1.32 & 23.95$\pm$0.44 & 13.16$\pm$0.91\\
				Reweight-R \cite{xia2019anchor} & 90.04$\pm$0.46 & 84.11$\pm$2.47 & 72.18$\pm$2.47 & 58.00$\pm$0.36 & 43.83$\pm$8.42 & 36.07$\pm$9.73\\
				Peer Loss \cite{liu2019peer} &89.12$\pm$0.76 & 83.26$\pm$0.42 & 74.53$\pm$1.22  & 61.16$\pm$0.64 & 47.23$\pm$1.23 & 31.71$\pm$2.06\\
				\SPL{} \cite{sieve2020} & 91.14$\pm$0.46 & 83.67$\pm$1.29 & 77.68$\pm$2.24 &66.47$\pm$0.45 & 58.99$\pm$1.49 & 38.55$\pm$3.25\\
				DivideMix\cite{Li2020DivideMix} & {93.33$\pm$0.14} & \textbf{95.07$\pm$0.11} & {85.50$\pm$0.71}  & {79.04$\pm$0.21} & {76.08$\pm$0.35} & {46.72$\pm$1.32}\\
				CAL\cite{zhu2021second} & {92.01$\pm$0.75} & {84.96$\pm$1.25} & {79.82$\pm$2.56}  & {69.11$\pm$0.46} & {63.17$\pm$1.40} & {43.58$\pm$3.30}\\
				
			    Ours & \textbf{93.68$\pm$0.12} & 94.97$\pm$0.09 & \textbf{94.95$\pm$0.11} & \textbf{79.61$\pm$0.19} & \textbf{76.58$\pm$0.25} & \textbf{59.40$\pm$0.46}\\
			   \hline
			\end{tabular}}
			}
		\end{center}
		\label{table:cifar-inst}
\end{table*}

\subsection{Datasets}

\noindent \textbf{Synthetic IDN Datasets.} Following previous studies on learning with IDN~\cite{zhu2021second}, our synthetic IDN datasets are created by adding two kinds of synthetic noise to CIFAR-10 and CIFAR-100 datasets~\cite{krizhevsky2009learning}, where CIFAR-10 contains 50,000 training images and 10,000 testing images from 10 different classes, CIFAR-100 contains 50,000 training images and 10,000 testing images from 100 classes.
Specifically, we use two kinds of synthetic IDN in our experiment:
\begin{itemize}
    \item Part-dependent label noise~\cite{xia2020parts}, which draws insights from human cognition that humans perceive instances by decomposing them into parts and estimates the IDN transition matrix of an instance as a combination of the transition matrices of different parts of the instance.
    \item Classification-based label noise~\cite{chen2020beyond}, which adds noise by i) collecting the predictions of each sample in every epoch during the training of a CNN classifier; ii) averaging the predictions and locate the class label with largest prediction probability other than the ground truth one for each instance as its noisy label; iii) flipping the labels of the samples whose largest probabilities falls in the top $r\%$ of all samples, where $r$ is a user-defined hyper-parameter.
\end{itemize}

\noindent \textbf{Real-world IDN Datasets.} 
Following~\cite{Li2020DivideMix}, we use Clothing1M~\cite{xiao2015learning} and Webvision 1.0~\cite{li2017webvision} to evaluate our method:
\begin{itemize}
    \item Clothing1M is a large scale dataset containing more than 1 million images of 14 kinds of clothes. As aforementioned, Clothing1M is highly imbalanced with its noise validated as IDN according to~\cite{chen2020beyond}. In our experiments, we use its noisy training set which contains 1 million images and report the performance on test set.
    \item Webvision is a large scale dataset which contains 2.4 million images from 1000 classes that are crawled from the web as ImageNet ILSVRC12 did. Following previous works~\cite{chen2019understanding,Li2020DivideMix}, we compare baseline methods on the first 50 classes of the Google image subset, and report the top-1 and top-5 performance on both Webvision validation set and ImageNet ILSVRC12.
\end{itemize}

\subsection{Implementation Details}

We follow DivideMix~\cite{Li2020DivideMix} and use MixMatch~\cite{berthelot2019mixmatch} for semi-supervised learning. For experiments on CIFAR-10 and CIFAR-100, we use ResNet-34~\cite{he2016deep} as the feature extractor following~\cite{zhu2021second}. We use similar hyperparameters to~\cite{Li2020DivideMix} across all 3 settings of CIFAR-10 and CIFAR-100 respectively. 
We train our model using a SGD optimizer with a momentum of 0.9 and a weight decay parameter of 0.0005. The learning rate is set as 0.02 in the first 150 epochs and reduced to 0.002 in the following 150 epochs. The warm up period is set as 10 epochs for CIFAR-10 and 15 epochs for CIFAR-100 respectively. 
For Clothing1M, we follow previous studies and use ImageNet pretrained ResNet-50 as the backbone. We train the model for 80 epochs. We set the learning rate as 0.002 in the beginning and reduce it to 0.0002 after 40 epochs of training. For Webvision 1.0, we follow~\cite{Li2020DivideMix} and use the Inception-Resnet v2\cite{szegedy2017inception} as the backbone. We train the model for 120 epochs. We set the learning rate as 0.01 in the first 50 epoch and 0.001 for the rest of the training. 

\begin{table}[!t]
	\caption{Classification accuracies on the (clean) test set of Clothing1M. Results of other method are copied from CAL\cite{zhu2021second}. Our method achieves state-of-the-art performance.}
	\begin{center}
	{
	\scalebox{0.8}{
		\begin{tabular}{c|c} 
			\hline 
			Method & Accuracy \\
			\hline \hline 
			CE (standard) &  68.94\\
			Forward $T$  \cite{patrini2017making} & 70.83\\
			Co-teaching \cite{han2018co}  &69.21 \\
			JoCoR \cite{wei2020combating} &70.30 \\
			$L_{\sf DMI}$ \cite{xu2019l_dmi} & 72.46\\
			PTD-R-V\cite{xia2020parts} & 71.67 \\
			DivideMix\cite{Li2020DivideMix} & 74.76 \\
			\SPL{} \cite{sieve2020} & 73.24\\
			CAL\cite{zhu2021second}  & 74.17 \\
			Ours & \textbf{75.40} \\
			\hline 
		\end{tabular}
		}
		}
	\end{center}
	\label{table:c1m}
\end{table}
\begin{table}[!t]
    \caption{Classification accuracies ($\%$) on CIFAR-10 with classification-based label noise of different noise ratios. Our method outperforms all previous ones in all settings.}
	\label{table_cifar10_idn}
	\begin{center}
	\scalebox{0.8}{
		\begin{tabular}{c|cccc}
			\hline
			Method	 &10$\%$ &20$\%$ &30$\%$ &40$\%$\\
			\hline \hline 
			\multirow{2}{*}{CE}			&91.25		&86.34		&80.87		&75.68 \\
			&$\pm$0.27  &$\pm$0.11  &$\pm$0.05  &$\pm$0.29\\
			\hline 
			\multirow{2}{*}{Forward\cite{patrini2017making}}	&91.06		&86.35		&78.87		&71.12 \\
			&$\pm$0.02  &$\pm$0.11  &$\pm$2.66  &$\pm$0.47\\
			\hline 
			\multirow{2}{*}{Co-teaching\cite{han2018co}}&91.22		&87.28		&84.33		&78.72 \\
			&$\pm$0.25  &$\pm$0.20  &$\pm$0.17  &$\pm$0.47\\
			\hline 
			\multirow{2}{*}{GCE\cite{zhang2018generalized}}		&90.97		&86.44		&81.54		&76.71 \\
			&$\pm$0.21  &$\pm$0.23  &$\pm$0.15  &$\pm$0.39\\
			\hline 
			\multirow{2}{*}{DAC\cite{thulasidasan2019combating}}		&90.94		&86.16		&80.88		&74.80 \\
			&$\pm$0.09  &$\pm$0.13  &$\pm$0.46  &$\pm$0.32\\
			\hline 
			\multirow{2}{*}{DMI\cite{xu2019l_dmi}}		&91.26		&86.57		&81.98		&77.81 \\
			&$\pm$0.06  &$\pm$0.16  &$\pm$0.57  &$\pm$0.85\\
			\hline 
			\multirow{2}{*}{SEAL\cite{chen2020beyond}}	    &91.32		&87.79		&85.30		&82.98 \\
			&$\pm$0.14  &$\pm$0.09  &$\pm$0.01  &$\pm$0.05\\				
			\hline 
			\multirow{2}{*}{Ours}	&\textbf{91.39}		&\textbf{88.36}		&\textbf{86.92}		&\textbf{84.18} \\
			&$\pm$0.08  &$\pm$0.11  &$\pm$0.68  &$\pm$0.40\\				
			\hline 
		\end{tabular}
	}
	\end{center}
\end{table}

\subsection{Experimental Results}

\noindent \textbf{CIFAR-10 and CIFAR-100}
As aforementioned, we evaluate our method on two kinds of IDN as follows:
\begin{itemize}
\item {\it Part-dependent label noise.} To facilitate a fair comparison, we borrow the noise used in CAL~\cite{zhu2021second} and follow CAL to test the performance of our method against 6 different settings, whose noise ratios vary between 0.2 and 0.6. 
As Table~\ref{table:cifar-inst} shows, our method outperforms previous methods in five in six settings, especially when the noise ratio and class number increase. For example, the improvement of CIFAR-100 with $\eta = 0.6$ is over 10\%.

\item {\it Classification-based label noise.} Following \cite{chen2020beyond}, we test our method against four different noise ratios, {10\%, 20\%, 30\% and 40\%}. To facilitate a fair comparison, we borrow the same noise from SEAL~\cite{chen2020beyond}. 
Note that compared to the aforementioned part-dependent label noise, the classification-based label noise used in this experiment is more challenging as it is generated by a CNN-based model.
As Table~\ref{table_cifar10_idn} shows, our method still outperforms previous methods in all four different settings. Similar as above, the improvement of our method becomes higher as the noise ratio increases, which demonstrates the effectiveness of our method under different kinds of IDNs.
\end{itemize}

\noindent \textbf{Clothing1M}
As aforementioned, Clothing1M contains over 1 million images from 14 classes collected from Internet, which makes it ideal to evaluate how different LNL methods perform against large-scale image datasets.
As Table~\ref{table:c1m} shows, our method outperforms all previous methods and achieves the state-of-the-art performance. Compared to DivideMix~\cite{Li2020DivideMix}, we further improve the accuracy by 0.64\%.

\noindent \textbf{Webvision and ImageNet ILSVRC12}
As Table~\ref{table:webvision} shows, our method achieves better performance on both top-1 and top-5 accuracy on ILSVRC12 and  Webvision. The higher improvement on ILSVRC12 suggests that our method is more robust to the domain difference and can generalize better.

\begin{table}[!t]
	\caption{Classification accuracies ($\%$) on (mini) Webvision and ILSVRC12.  Numbers
    denote top-1 (top-5) accuracy (\%) on the WebVision and the ImageNet ILSVRC12 validation sets.}
    \label{table:webvision}
    \begin{center}
    \scalebox{0.8}{
	\begin{tabular}	{l|c|c|c|c}
		\hline 
			\multirow{2}{*}{Method}  & \multicolumn{2}{c|}{WebVision} & \multicolumn{2}{c}{ILSVRC12}\\
			\cmidrule{2-5}
			& top1 & top5& top1 & top5\\
			\hline \hline 	
			F-correction~\cite{Giorgio_CVPR_2017} & 61.12 & 82.68&  57.36 &82.36\\		
		    
			Decoupling~\cite{decouple} & 62.54 &84.74& 58.26 &82.26\\
				D2L~\cite{Ma_ICML_2018} &62.68 &84.00&  57.80 &81.36\\
			MentorNet~\cite{jiang2018mentornet}  &63.00 &81.40&  57.80 &79.92\\	
			
			Co-teaching~\cite{co-teaching}&63.58 &85.20&   61.48 &84.70\\			
			Iterative-CV~\cite{chen2019understanding} 	&  65.24 &85.34&  61.60 &84.98\\	
			DivideMix\cite{Li2020DivideMix} &77.32 &91.64& 75.20 &90.84 \\
			NGC\cite{Li2020DivideMix} &79.16 &91.84& 74.44 &91.04 \\
			
			\hline
			Ours & \textbf{79.36} &\textbf{93.64}& \textbf{76.08} &\textbf{93.86}\\
		\hline 
	\end{tabular}}
    \end{center}
\end{table}		

\begin{table}[!t]
\caption{Ablation study on our Feature Clustering (Stage 1) and Consistency Classification (Stage 2). The models with neither stages are trained with cross-entropy loss (\ie CE baseline).}
\label{table:ablation}
	\begin{center}
	{
		\begin{tabular}{c|cc|c}
    		\hline
            Dataset & \begin{tabular}[c]{@{}l@{}}Feature\\ Clustering\end{tabular} & \begin{tabular}[c]{@{}l@{}}Consistency\\ Classification\end{tabular} & Accuracy \\
            \hline \hline
            \multirow{3}{*}{\begin{tabular}[c]{@{}l@{}}CIFAR-100\\ ($\mu$=0.6)\end{tabular}} & & & 25.68 \\
            & \checkmark & & 53.60 \\
            
            & & \checkmark & 51.41 \\
            & \checkmark &  \checkmark & 59.40 \\
            \hline
            
            \multirow{3}{*}{Clothing1M} & & & 68.94 \\
            & \checkmark & & 73.32 \\
            
            & & \checkmark & 74.26 \\
            & \checkmark & \checkmark & 75.40 \\
            \hline
        \end{tabular}
	}
	\end{center}
\end{table}

\subsection{Ablation Study}

We conduct an ablation study on the two stages of our method.
Specifically, we provide the performance of our method on both CIFAR-100, a synthetic IDN dataset with noise ratio $\eta = 0 .6$ and Clothing1M, a highly-imbalanced dataset with real-world IDN. 
We also compare our method to standard CE baseline (\ie neither stages are applied). As Table~\ref{table:ablation} shows, our method benefits from each stage in terms of the performance on both datasets, and achieves the best results when both stages are employed.

\subsection{Performance against Class Imbalance}

\begin{figure}[t]
  \centering
   \includegraphics[width=0.4\linewidth]{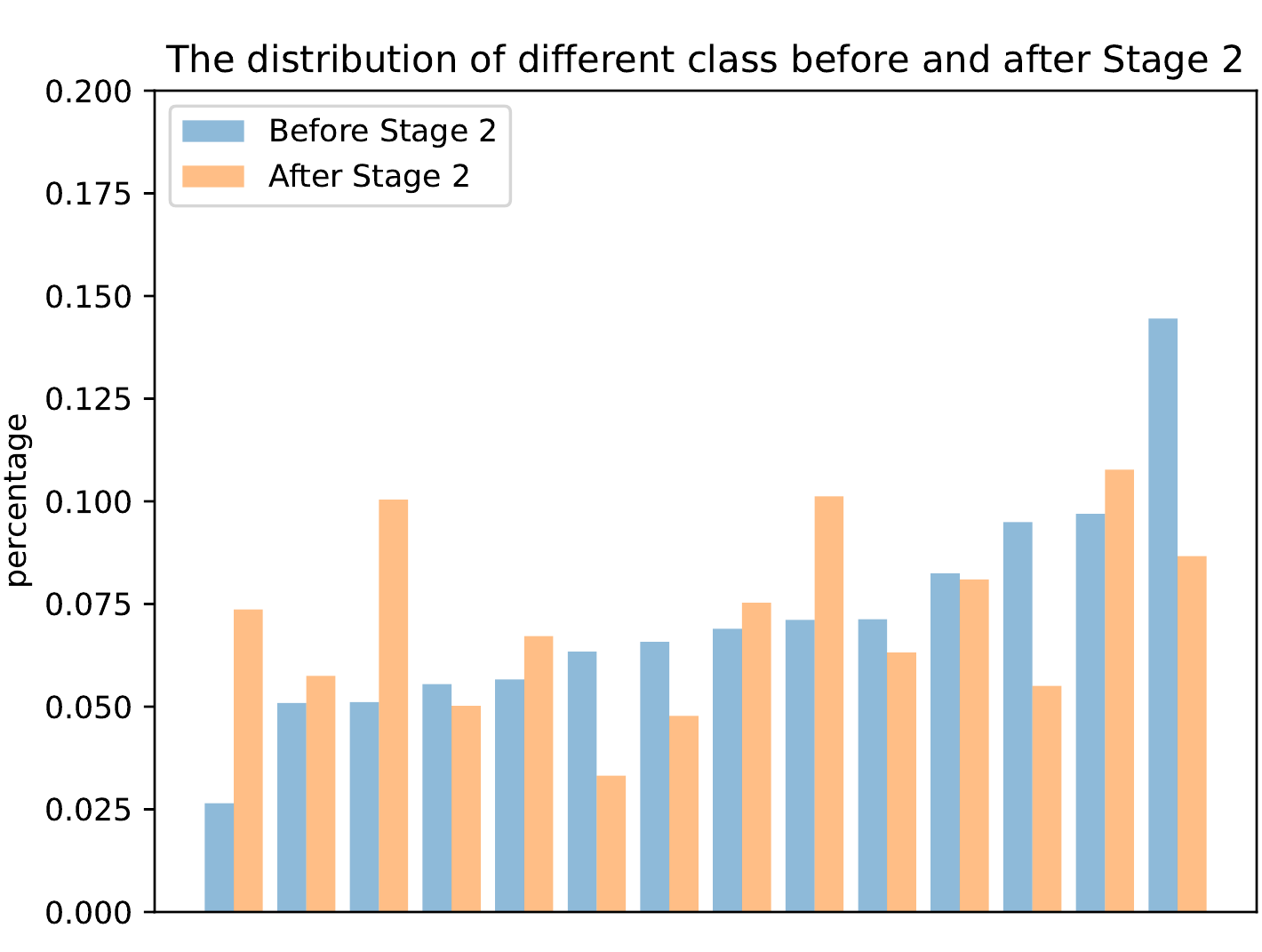}

   \caption{The distributions of different classes in the validation set of Clothing1M before and after the consistency-based classification (Stage 2). After our consistency-based classification, the distribution becomes more balanced.}
   \label{fig:clothing_distri}
\end{figure}

We select the highly-imbalanced Clothing1M to test the performance of our method against class imbalance. 
Specifically, we are concerned on the distribution~(proportion of class-wise sample number w.r.t the whole dataset) changes of all 14 classes within our selected clean samples 
before and after our consistency-based classification. 
Since Clothing1M does not contain the ground truth labels for its noisy training set, we mix some samples from its validation set that contains both clean and noisy labels with the original noisy training set, and report the distributions of the validation samples. As Fig.~\ref{fig:clothing_distri} shows, the percentages of most of the rare classes increase after our consistency-based classification, while the percentages of the rich classes decrease. In addition, we observed biggest changes occur in the rarest and richest classes.

\begin{figure}[t]
  \centering
   \includegraphics[width=0.4\linewidth]{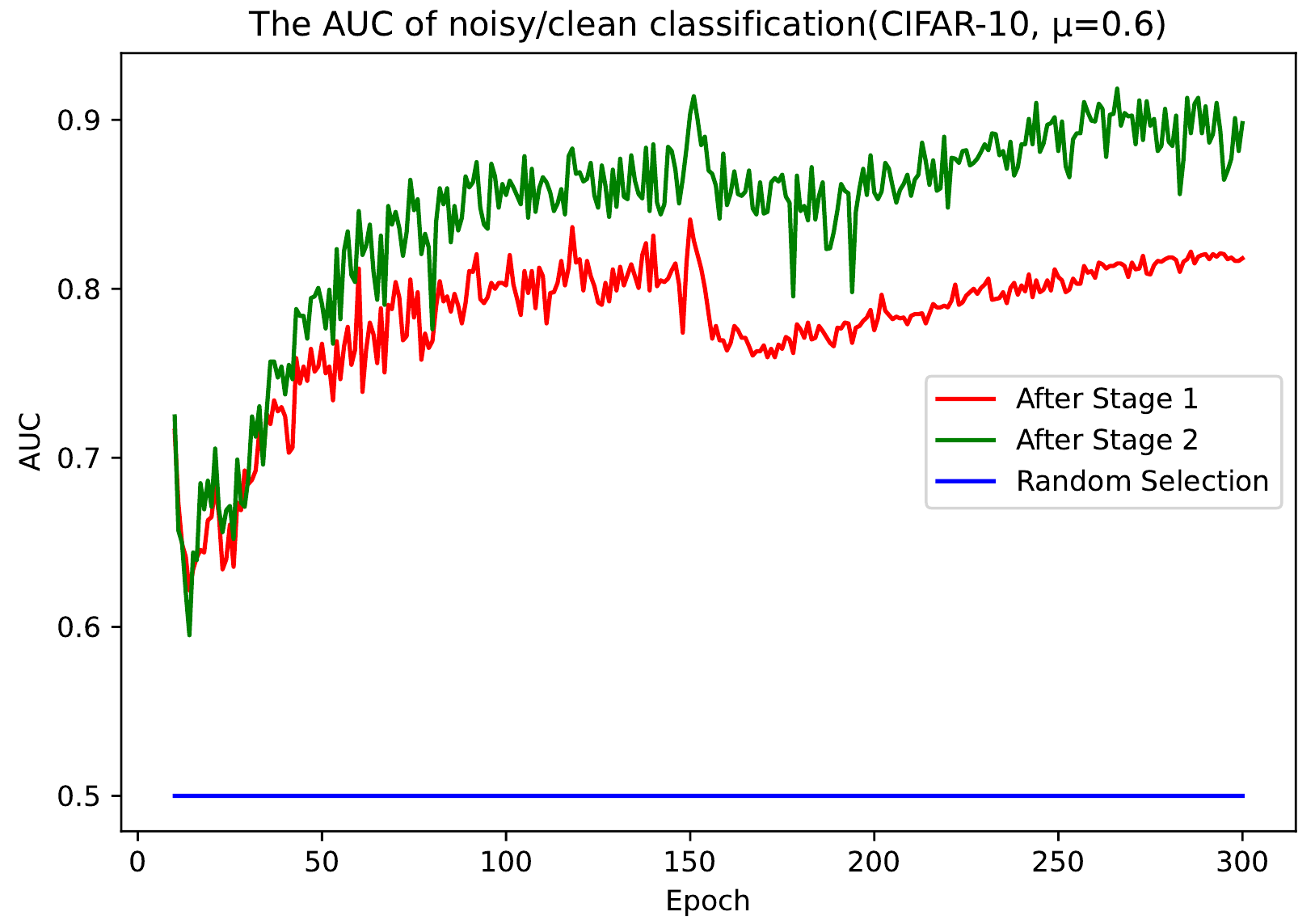}

   \caption{The AUC of noisy vs. clean classification of our method. The second stage steadily improve the AUC of classification. The performance drop at 150 epoch is due to a learning rate change.}
   \label{fig:auc_cifar}
\end{figure}

\begin{figure}[t]
  \centering
  \hspace*{\fill}%
    \subfloat[]{                    
     \includegraphics[width=0.3\linewidth]{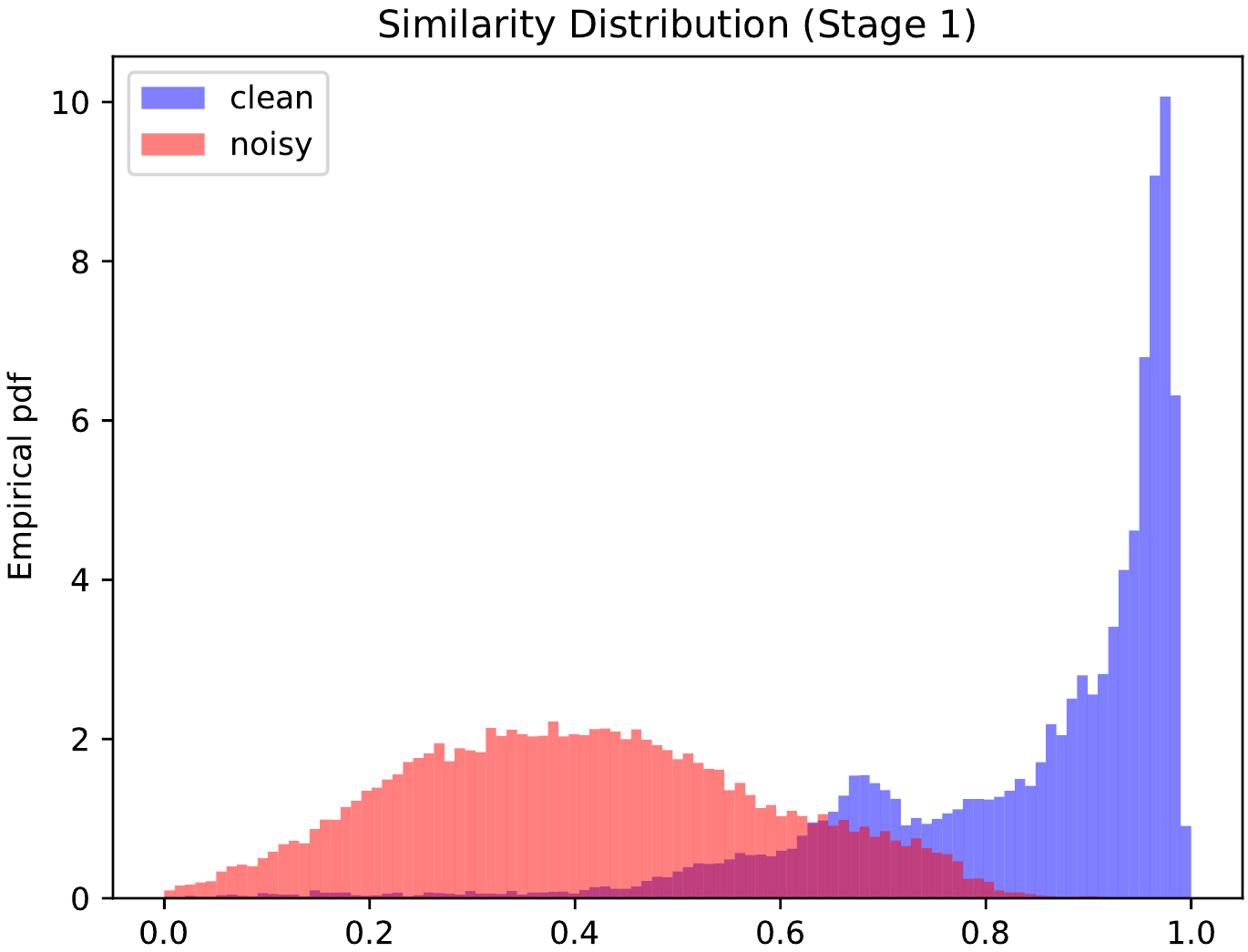}
    }
    \hfill%
    \subfloat[]{%
        \includegraphics[width=0.3\linewidth]{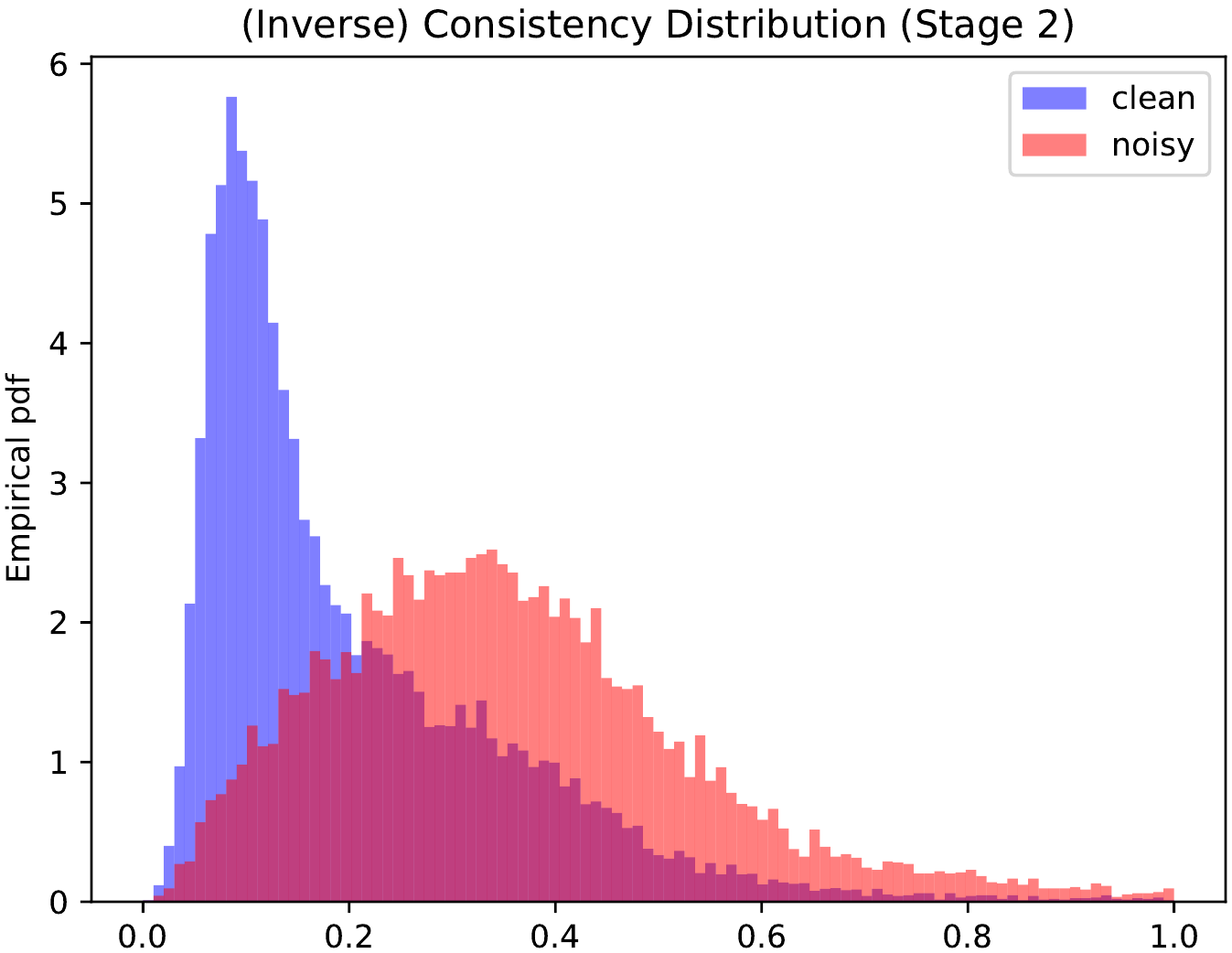}
    }%
    \hspace*{\fill}%
   \caption{The probability distribution function of clean/noisy samples respectively for CIFAR-10 ($\mu$=0.6). The range of statistics is normalized to 0 to 1. (a) The similarity distribution of stage 1. (b) The (inverse) consistency distribution of stage 2.}
   \label{fig:pdf_cifar}
\end{figure}

\subsection{AUC of Noisy vs. Clean Classification}

Given the prediction probabilities of stage 1 and stage 2, we calculate the area under curve (AUC) of our noisy vs. clean classification on CIFAR-10 with a noise ratio of 0.6. 
As Fig.~\ref{fig:auc_cifar} shows, compared to the performance of random selection, both stages of our method can improve the AUC of classification, and the second stage further improve the AUC over the first stage. In addition, it can be observed that the accuracy of noisy vs. clean is improved as the training progresses. The performance decrease occurred around 150 epoch is due to a 0.1-fold decrease of the learning rate.
Beside, we provide the probability distribution function of similarity and consistency in Fig.~\ref{fig:pdf_cifar}. Both metrics are effective in distinguishing clean and noisy samples.

\section{Conclusion}

In this paper, we propose a two-stage method to address the problem of learning with instance-dependent noisy labels in the presence of inter-class imbalance problem. 
In the first stage, we identify ``easy'' clean samples that are close to the class-wise prediction centers using a class-level feature clustering procedure. We also address the class imbalance problem by augmenting the clustering with an entropy-based rare class aggregation technique. In the second stage, we further identify the remaining ``difficult'' clean samples that are close to the ground truth class boundary based on the consistency of two classifier heads.
We conducted extensive experiments on several challenging benchmarks to demonstrate the effectiveness of the proposed method. 

\section*{Acknowledgements}This work was supported in part by the Guangdong Basic and Applied Basic Research Foundation (No.2020B1515020048), in part by the National Natural Science Foundation of China (No.61976250, No.U1811463), in part by the Hong Kong Research Grants Council through Research Impact Fund (Grant R-5001-18), and in part by the Guangzhou Science and technology project (No.202102020633).

\clearpage
%
%
\bibliographystyle{splncs04}
\bibliography{egbib}
\end{document}